\documentclass[10pt,twocolumn,letterpaper]{article}

\usepackage[pagenumbers]{wacv} 

\usepackage{tikz}
\usepackage{pifont}
\usepackage{xcolor}
\usepackage{multirow}
\usepackage[table]{xcolor}
\usepackage{threeparttable}
\usetikzlibrary{calc}

\definecolor{wacvblue}{rgb}{0.21,0.49,0.74}
\usepackage[pagebackref,breaklinks,colorlinks,allcolors=wacvblue]{hyperref}

\title{Learning to Reason Like a Cytopathologist:\\
Clinically Grounded Multiscale Evidential Learning for Cervical Cytology}

\author{
Ajeet Kumar Singh$^{1,2}$ \quad Nehal Mishra$^{2}$ \quad Anurag Tripathi$^{2}$\\
Santanu Chaudhury$^{1}$ \quad Dorien Herremans$^{3}$\\[6pt]
{\normalsize $^{1}$IIT Delhi, India \quad $^{2}$Info Origin INC. \quad $^{3}$Singapore University of Technology and Design}\\[4pt]
{\small $^{1}$\tt srz258064@iitd.ac.in, santanu.choudhury@gmail.com}\\
{\small $^{2}$\tt \{nehal.mishra, anurag.tripathi\}@infoorigin.com} \quad
{\small $^{3}$\tt dorien\_herremans@sutd.edu.sg}
}

\begin{document}

\maketitle

\begin{abstract}
Cervical cytology presents a fundamentally ambiguous diagnostic landscape, where equivocal categories such as Atypical Squamous Cells of Undetermined Significance (ASC-US) occupy a morphological grey zone that is definitional. For clinical decision support, the goal extends beyond assigning a definitive label to every sample: cases must also be distinguished by whether the available evidence supports a confident decision or warrants further consideration, a distinction conventional classification models are not designed to explicitly represent. We propose a multiscale evidential framework that models cervical cytology along an ordinal disease continuum using clinically grounded soft anchors, representing progression from benign reactive changes to precancerous lesions and malignancy. Three biologically motivated spatial contexts—nuclear morphology, cytoplasmic characteristics, and neighbourhood cellular architecture—are independently encoded using a shared frozen DINOv2 backbone, preserving inter-scale disagreement that early fusion may obscure. Calibrated predictions are transformed into Dempster--Shafer belief masses and fused with reliability discounting, allowing conflicting evidence and insufficient support to remain explicitly uncommitted rather than forcing a decision. The fused belief state drives interpretable routing that separates confidently classified cases from those with irreducible evidential insufficiency, enabling clinically meaningful triage. Across two independent publicly available conventional cytology datasets, RIVA and CRIC Cervix, the framework generalises consistently, with uncertainty strongly associated with diagnostic difficulty: incorrect predictions carry 1.94$\times$--4.13$\times$ higher uncertainty than correct ones.
\end{abstract}

\section{Introduction}
\label{sec:intro}

Cervical cancer is among the most prevalent yet preventable cancers
worldwide, with the Pap smear serving as a primary screening modality~\cite{alrajjal2021}.
Cytological findings are reported according to The Bethesda System (TBS),
the internationally adopted standard updated in 2014, from which eight
diagnostic categories are considered in this work~\cite{nayar2015book,nayar2015paper}
(Table~\ref{tab:tbs}). Importantly, these categories are not equally
definitive: TBS distinguishes definitive interpretations (NILM, ENDO, INFL LSIL, HSIL, SCC) from equivocal grey-zone categories (ASC-US, ASC-H). This
distinction reflects the fact that some cellular morphologies cannot be
reliably separated at the single-cell level.

\begin{table}[t]
\centering
\caption{The eight TBS diagnostic categories considered in this work.}
\label{tab:tbs}
\small
\setlength{\tabcolsep}{10pt}
\renewcommand{\arraystretch}{1.05}
\begin{tabular}{@{}l|c|c@{}}
\toprule
\textbf{Class} & \textbf{Clinical position} & \textbf{Interpretation} \\
\midrule
NILM   & Benign & Definitive \\
ENDO   & Benign & Definitive \\
INFL   & Benign & Definitive \\
ASC-US & Lesion & Equivocal \\
LSIL   & Lesion & Definitive \\
ASC-H  & Lesion & Equivocal \\
HSIL   & Lesion & Definitive \\
SCC    & Lesion & Definitive \\
\bottomrule
\end{tabular}
\end{table}

This structure also reflects an underlying ordinal progression in
clinical severity. Persistent high-risk HPV infection can drive
progressive cellular changes, with TBS categories occupying different
positions along this trajectory and carrying different clinical risks.
In particular, the risk of CIN3+ -- a surrogate measure of high-grade cervical disease and cancer risk, increases substantially across the spectrum, from near-zero for NILM to 60--100\% for HSIL, with ASC-US and LSIL occupying intermediate positions~\cite{egemen2020}. Thus, the categories are clinically ordered rather than a collection of independent labels. At the same time, important category boundaries remain morphologically ambiguous:
INFL and ASC-US can exhibit overlapping nuclear features, while ENDO
cells are benign but can exhibit morphology that mimics HSIL
~\cite{nayar2015paper}. These cases illustrate why a visually plausible
prediction is not necessarily a definitive one.

Cytopathologists address such ambiguity by integrating evidence across
spatial scales, from cellular grouping and neighbourhood context at
lower magnification to nuclear chromatin and membrane detail at higher
magnification~\cite{alrajjal2021}. For automated analysis, this creates
three coupled challenges: distinguishing visually similar categories
whose relationships are ordinal rather than independent~\cite{cao2020rank,shi2023corn};
handling genuinely ambiguous class boundaries; and combining
complementary evidence observed at different spatial scales. A
classifier operating on an isolated nucleus patch therefore captures
only part of the available diagnostic evidence. Moreover, conventional
multiscale architectures typically fuse features before prediction,
which can suppress disagreement between scales precisely when that
disagreement may indicate an ambiguous case. Hard categorical
predictions further provide no explicit representation of insufficient
or conflicting evidence.

We propose a framework that treats these properties of cervical
cytology as design constraints rather than incidental characteristics.
Specifically, we introduce: \textbf{(i)} clinically grounded, non-uniform ordinal soft anchors that encode the NILM-to-SCC progression and provide a
structured representation of class relationships; \textbf{(ii)} three
spatially independent scale-specific MLPs operating on frozen DINOv2
features, allowing evidence from different spatial scales to remain
separable and preserving their disagreement for subsequent fusion;
\textbf{(iii)} a Dempster--Shafer fusion stage that combines calibrated
scale-specific likelihoods into explicit belief masses
($m_{\mathrm{benign}}$, $m_{\mathrm{lesion}}$, and
$m_{\mathrm{uncertain}}$), allowing evidentially insufficient cases to
be routed for further review rather than forcing a categorical
decision; and \textbf{(iv)} cross-dataset evaluation on RIVA~\cite{riva2025}
and CRIC Cervix~\cite{rezende2021}, showing that incorrect predictions
exhibit systematically greater uncertainty than correct predictions
and that cases assigned high uncertainty are concentrated around the
equivocal categories identified by TBS.

\section{Related Work}
\label{sec:related_works}

Deep learning approaches to cervical cytology have achieved competitive performance on isolated-cell benchmarks~\cite{diniz2021,khan2023,liu2022,diniz_hierarchical2021},
but studies differ substantially in diagnostic label groupings, dataset usage, and evaluation protocols, complicating direct comparison across reported results~\cite{martinez2025,robust2026}. For RIVA, a recently introduced conventional-smear benchmark, the literature is still emerging, with the 2026 ISBI challenge establishing an eight-category detection-and-classification benchmark~\cite{riva_challenge2026,kogan2026detection, kang2026riva}. A systematic review of 77 studies similarly identifies limited generalizability and insufficient representation of real-world cytological complexity as persistent challenges~\cite{valles2026}. Multiscale architectures have been explored to incorporate complementary morphological information across spatial contexts~\cite{khan2023,liu2022}, however these predictions are typically used to improve classification rather than to explicitly characterize the evidence provided by each spatial view.

Uncertainty-aware and evidential learning provides a framework for
representing predictions beyond a single categorical decision.
Evidential deep learning models uncertainty through learned evidence
\cite{sensoy2018}, while Dempster--Shafer theory (DST)
\cite{shafer1976,dempster1967} represents support for competing hypotheses together with uncommitted evidence. Recent work has extended DS-based reasoning to reliable classification~\cite{kempkes2024}, set-valued evidential prediction~\cite{bezirganyan2025emsec}, class inference \cite{shiraishi2026}, and dynamic evidence learning~\cite{fan2025}.
These developments establish DST as a general framework for reasoning with incomplete or conflicting evidence, but its use as an explicit fusion mechanism for complementary spatial predictions in cervical cytology remains unexplored.

Our work brings these two directions together by treating the nuclear, cellular, and neighbourhood views of a cell as separate evidence sources. Rather than fusing their features or predictions into a single categorical decision, we retain their evidential states until late fusion, allowing agreement, conflict, and insufficient evidence to influence the final decision. We therefore use established DST machinery to formulate an ambiguity-aware multiscale classification and routing framework tailored
to cervical cytology.
\section{Method}

The preceding sections identify three requirements for reliable cervical cytology classification: spatial contexts provide complementary morphological evidence, the Bethesda categories follow an ordinal continuum, and some cells remain equivocal even under multiscale examination. We therefore formulate the task as multiscale evidential inference, keeping predictions from different spatial views independent until evidence is explicitly combined.

\begin{figure*}[t]
    \centering
    \includegraphics[width=\textwidth]{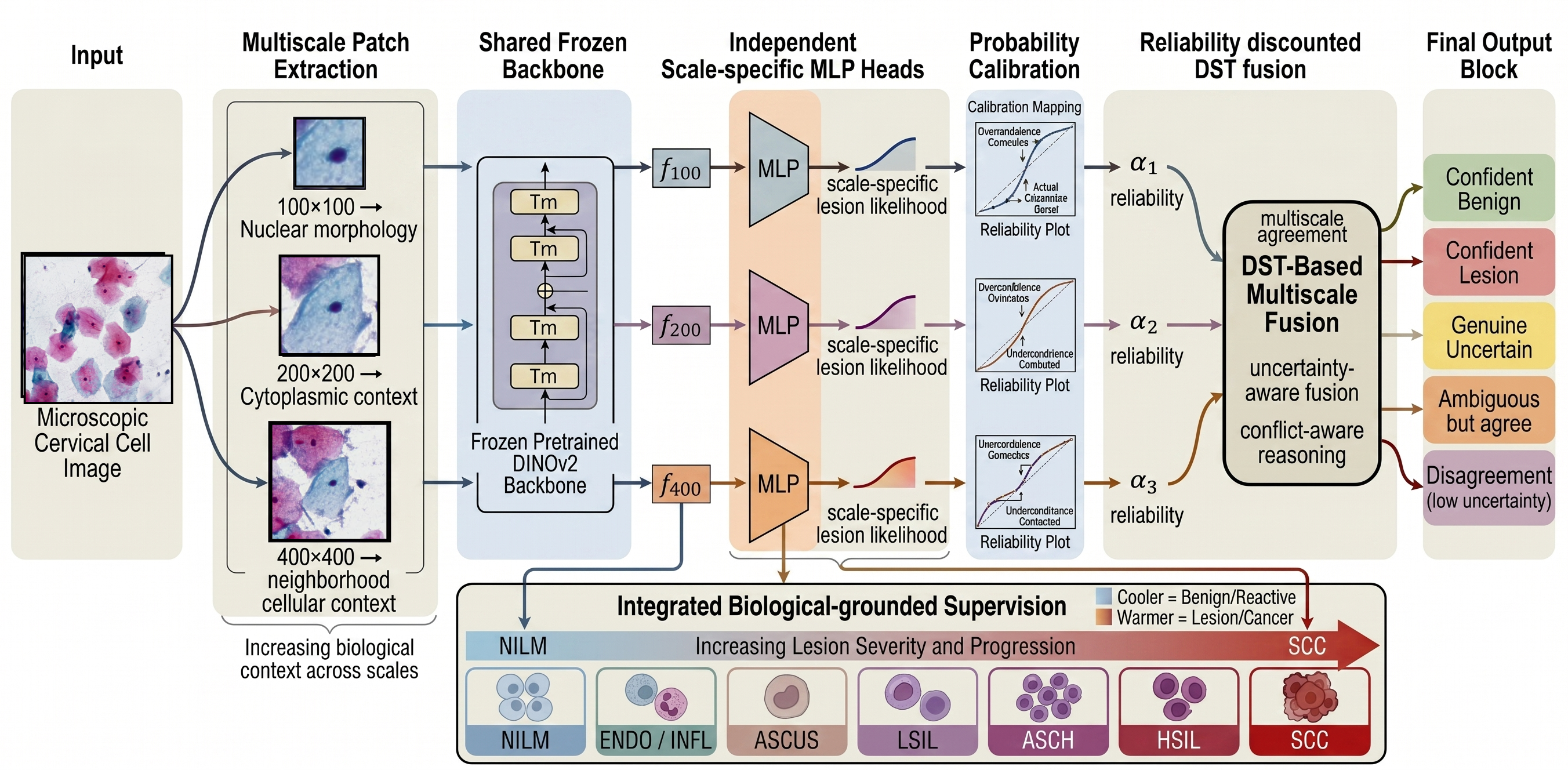}
    \caption{Overview of the proposed framework. Multiscale patches capturing nuclear, cytoplasmic, and neighbourhood context are processed through a shared frozen DINOv2 backbone and independent scale-specific heads. Calibrated predictions are transformed into Dempster--Shafer evidence and fused with reliability discounting to support explicit uncertainty for ambiguity-aware routing.}
    \label{fig:overview}
\end{figure*}

As illustrated in Fig.~\ref{fig:overview}, each annotated cell is represented using three nucleus-centred patches of 100, 200, and 400 pixels. The 100-pixel crop follows the cell annotation convention provided by RIVA; the 200- and 400-pixel crops were defined by us to progressively incorporate cytoplasmic and neighbourhood context. For CRIC Cervix, where cells are specified by nucleus coordinates, we apply the same multiscale cropping principle around the annotated cell. Thus, the same spatial hierarchy is maintained across both datasets.

Each crop is encoded by a shared frozen DINOv2 backbone, followed by an independent prediction head for each scale. The heads are trained using ordinal soft anchors for all eight Bethesda categories, producing a scalar lesion-transition score rather than a conventional eight-class probability vector. These scores are subsequently calibrated and interpreted within a binary benign--lesion evidential frame. Each scale therefore contributes evidence that can support either hypothesis or remain uncommitted. The three evidential sources are reliability-discounted and fused using Dempster--Shafer theory (DST), producing a belief state that is finally used to route the cell to an appropriate downstream action.

\subsection{Multiscale Feature Extraction and Ordinal Prediction}

For each annotated cell, we extract the three nucleus-centred patches described above. Each view is encoded independently using a shared frozen DINOv2 ViT-B/14 backbone~\cite{oquab2023dinov2}. DINOv2's self-supervised pretraining on diverse visual data provides representations of texture, shape, and relational structure, reducing reliance on explicitly engineered cytological measurements such as segmentation-dependent N/C ratio or chromatin grading. The backbone is not finetuned to reduce overfitting to dataset-specific staining and acquisition characteristics, while the subsequent scale-specific heads learn the information associated with each spatial context.

A separate MLP prediction head is trained for each scale so that the three views retain independent predictions for subsequent evidential fusion. Standard binary supervision would assign all benign categories the same target and all lesion categories another, discarding the ordered progression between clinically distinct Bethesda categories. We instead retain all eight categories during supervision by assigning each class a scalar soft anchor $y_c\in[0,1]$ representing its ordinal position along the benign-to-malignancy continuum and keeping the equivocality of ASC-US and ASC-H intact.

Each MLP is trained with BCEWithLogitsLoss against its class-specific soft anchor, producing a continuous lesion-transition score whose value reflects the ordinal position of the TBS category rather than a hard benign--lesion label. The anchors are informed by the CIN3+ risk progression described by the ASCCP risk-based framework, but represent ordinal positions rather than raw clinical risk probabilities~\cite{egemen2020}. The binary evidential frame therefore interprets, rather than replaces, the learned ordinal prediction for final benign--lesion routing.

\begin{table}[t]
\centering
\footnotesize
\setlength{\tabcolsep}{4pt}
\renewcommand{\arraystretch}{0.92}
\caption{Clinically grounded, non-uniform ordinal soft anchors for the eight TBS categories. The ordering and relative spacing encode clinically motivated proximity along the benign-to-malignant continuum; anchor values are ordinal positions rather than raw clinical risk probabilities.}
\label{tab:anchors}
\begin{tabular}{l l l c}
\toprule
\textbf{Class} & \textbf{Clinical position} & \textbf{CIN3+ risk} & $\mathbf{y_c}$ \\
\midrule
NILM   & Clearly benign                 & Negligible             & 0.00 \\
ENDO   & Benign; mimics HSIL           & Negligible             & 0.05 \\
INFL   & Reactive; overlaps ASC-US     & Very low               & 0.25 \\
ASC-US & Definitionally equivocal       & Low                    & 0.60 \\
LSIL   & Low-grade lesion               & Moderate               & 0.85 \\
ASC-H  & Cannot exclude HSIL            & High                   & 0.90 \\
HSIL   & High-grade lesion              & Very high              & 0.95 \\
SCC    & Definite malignancy             & Confirmed malignancy   & 1.00 \\
\bottomrule
\end{tabular}
\end{table}

The non-uniform anchor spacing encodes clinically motivated proximity along the TBS continuum: benign categories occupy a compact region, while the larger transitions toward ASC-US and LSIL distinguish equivocal from definitive lesion states, and the closely spaced LSIL--ASC-H--HSIL--SCC categories reflect their related but progressively severe lesion spectrum~\cite{alrajjal2021}. In particular, the ASC-US anchor places this definitionally equivocal category between the benign and lesion regions without collapsing it into either, allowing the scalar output to retain graded ordinal information for subsequent evidential fusion.

\subsection{Calibrated Predictions and Evidential Fusion}

The ordinally trained MLP produces, for each spatial scale $s$, a scalar logit $z_s$. We first convert this output to a lesion probability:
\begin{equation}
p_s=\sigma(z_s)=\frac{1}{1+\exp(-z_s)}.
\label{eq:sigmoid}
\end{equation}

Although the MLP was trained using the eight ordinal targets, the evidential stage operates on the binary hypotheses $\Theta=\{B,L\}$, where NILM, ENDO, and INFL form the benign group and ASC-US, LSIL, ASC-H, HSIL, and SCC form the lesion group. This binary frame is used only to express the routing evidence; the ordinal
information has already been incorporated into the learned scalar transition score.

Because neural networks may produce poorly calibrated, systematically overconfident or underconfident probabilities, and these probabilities directly determine the amount of evidence assigned to each hypothesis, calibration is performed independently for each scale before evidence construction. We evaluate temperature scaling~\cite{guo2017calibration} and 5-fold cross-validated isotonic regression ~\cite{niculescu2005predicting,zadrozny2002calibration}, selecting the strategy for each dataset based on validation performance. A scale-specific decision threshold $t_s$ is then selected on the validation set by maximizing binary macro-F1 over candidate thresholds
in $[0.1,0.9]$. This threshold defines the benign-lesion decision boundary used by the evidential stage.

The calibrated probability $p_s^{\mathrm{cal}}$ is next normalized relative to this scale-specific decision boundary. Because the three spatial views may have different operating thresholds, their raw probabilities are not directly comparable as evidence. We therefore map the decision threshold to zero and the two probability endpoints to $-1$ and $+1$:
\begin{equation}
p^{\mathrm{norm}}_s=
\begin{cases}
\dfrac{p_s^{\mathrm{cal}}-t_s}{1-t_s}, & p_s^{\mathrm{cal}}\geq t_s,\\[5pt]
\dfrac{p_s^{\mathrm{cal}}-t_s}{t_s}, & p_s^{\mathrm{cal}}<t_s.
\end{cases}
\label{eq:norm}
\end{equation}
Thus, $p_s^{\mathrm{norm}}=0$ corresponds to the decision boundary, negative values indicate benign-oriented evidence, and positive values indicate lesion-oriented evidence. Its magnitude measures the prediction's distance from the boundary on a common scale across spatial views.

We convert this normalized signed distance into an evidence strength:
\begin{equation}
e_s=1-\exp\left(-\lambda(p^{\mathrm{norm}}_s)^2\right),
\label{eq:evidence}
\end{equation}
where $\lambda$ is the \textbf{evidence-sharpness parameter}, controlling how rapidly evidence strength increases as the prediction moves away from the decision boundary. Predictions close to the boundary therefore provide little committed evidence, whereas predictions farther from it provide stronger evidence.

The evidence strength is converted into a \textbf{basic probability
assignment (BPA)} over the frame of discernment $\Theta$:
\begin{equation}
\begin{aligned}
m^s_B &= e_s\mathbf{1}[p^{\mathrm{norm}}_s\leq0], &
m^s_L &= e_s\mathbf{1}[p^{\mathrm{norm}}_s>0],\\
m^s_U &= 1-e_s.
\end{aligned}
\label{eq:bpa}
\end{equation}
Here, $\mathbf{1}[\cdot]$ denotes the indicator function.
$m^s_B$ and $m^s_L$ represent mass assigned to the benign and lesion hypotheses, respectively, while $m^s_U$ represents uncertainty mass when the evidence is insufficient to commit to either hypothesis. Together, these three quantities form the scale-specific BPA.

The three spatial views may have different reliability when their predictions conflict. We therefore estimate a \textbf{scale reliability factor $\alpha_s$} from validation accuracy on samples where the scale-specific binary votes disagree, and map these accuracies to
$[0.40,1.00]$:
\begin{equation}
\alpha_s=0.40+0.60
\frac{\operatorname{acc}^{\mathrm{conflict}}_s
-\min_i\operatorname{acc}^{\mathrm{conflict}}_i}
{\max_i\operatorname{acc}^{\mathrm{conflict}}_i
-\min_i\operatorname{acc}^{\mathrm{conflict}}_i}.
\label{eq:alpha}
\end{equation}
Reliability discounting reduces the committed evidence of a less
reliable source and transfers the residual mass to the uncertainty
state:
\begin{equation}
\tilde m^s_B=\alpha_s m^s_B,\quad
\tilde m^s_L=\alpha_s m^s_L,\quad
\tilde m^s_U=\alpha_s m^s_U+(1-\alpha_s).
\label{eq:discount}
\end{equation}
The reliability factors fixed during validation are held constant at test time.

Finally, the three discounted BPAs are fused sequentially using
Dempster's rule. For two sources, let $\tilde{\mathbf m}^{1}$ and
$\tilde{\mathbf m}^{2}$ denote any two discounted BPAs. Their conflict
mass is
\begin{equation}
K=\tilde m^1_B\tilde m^2_L+\tilde m^1_L\tilde m^2_B.
\label{eq:conflict}
\end{equation}
The resulting combined BPA is
\begin{equation}
m_B=
\frac{\tilde m^1_B\tilde m^2_B+\tilde m^1_B\tilde m^2_U+
\tilde m^1_U\tilde m^2_B}{1-K},
\label{eq:mb}
\end{equation}
\begin{equation}
m_L=
\frac{\tilde m^1_L\tilde m^2_L+\tilde m^1_L\tilde m^2_U+
\tilde m^1_U\tilde m^2_L}{1-K},
\label{eq:ml}
\end{equation}
\begin{equation}
m_U=
\frac{\tilde m^1_U\tilde m^2_U}{1-K}.
\label{eq:mu}
\end{equation}
Here, $K$ denotes the mass arising from conflicting benign and lesion evidence. For the three spatial scales, this two-source rule is applied twice: the result of combining the first two scales is subsequently
combined with the discounted BPA from the third scale. Agreement between sources increases support for the corresponding hypothesis, while conflicting evidence contributes to $K$ and insufficient evidence remains represented by $m_U$. The resulting $(m_B,m_L,m_U)$ forms the fused BPA used by the routing policy below.

\subsection{Evidence-Based Routing}

The final belief state $(m_B,m_L,m_U)$ is used together with the three scale-specific votes to determine the routing action. A scale votes benign when $p_s^{\mathrm{cal}}\leq t_s$ and lesion otherwise.
The routing policy distinguishes confident agreement, high evidential insufficiency, unanimous predictions lacking sufficient evidence, and disagreements that can be resolved by the fused evidence.

\begin{table}[t]
\centering
\footnotesize
\setlength{\tabcolsep}{3pt}
\renewcommand{\arraystretch}{0.9}
\caption{Five-way evidence-based routing. $m_U$ denotes the fused
uncertainty mass.}
\label{tab:routing}
\begin{tabular}{l l l}
\toprule
\textbf{Category} & \textbf{Action} & \textbf{Condition} \\
\midrule
Confident Benign
& Benign sample
& All 3 votes benign, $m_U\leq0.40$ \\
Confident Lesion
& Lesion sample
& All 3 votes lesion, $m_U\leq0.40$ \\
Genuine Uncertain
& Further review
& Split vote, $m_U>0.50$ \\
Ambiguous (agree)
& Further review
& Unanimous vote, $m_U>0.40$ \\
Disagree (confident)
& DST-resolved
& Split vote, $m_U\leq0.50$ \\
\bottomrule
\end{tabular}
\end{table}

Samples with $m_U>0.50$ form the \textbf{uncertain pool}, which is further distinguished by whether the three scale-specific votes agree or disagree. The routing therefore does not treat high $m_U$ as a third
diagnosis; rather, it identifies samples for which the available multiscale evidence does not support a confident single-branch decision. Conversely, when the scale votes disagree but $m_U$ remains low, the fused belief state provides a principled resolution rather
than relying on majority vote.

For CRIC Cervix, we additionally apply a dataset-specific NILM vote filter to the uncertain pool to suppress spurious lesion votes. Samples with fewer than two of three lesion votes are reassigned to the benign branch. This adjustment is not part of the core evidential formulation and is discussed in section \ref{sec:limitations} and the supplementary material.

The resulting framework is therefore a fixed evidential formulation rather than a collection of independently tuned components. The spatial scales and ordinal anchors establish the clinical formulation, with patch sizes dictated by the available annotation conventions, while evidential fusion and routing provide a consistent procedure for combining and interpreting the resulting predictions. Only calibration strategy and scale-specific decision thresholds are selected on validation data to account for dataset-specific differences in acquisition and class distribution. Once selected, all parameters remain fixed for test evaluation.
\section{Experiments}
\label{sec:experiments}

\subsection{Datasets}
\label{sec:datasets}

\textbf{RIVA.}
RIVA~\cite{riva2025} is a large-scale conventional Pap smear dataset
covering all eight TBS categories, with cell annotations reviewed by up
to four cytopathologists. We use train/validation/test split of 10,613, 2,654, and 2,682 annotated cells, respectively. RIVA includes both INFL and ASC-US, providing a challenging benign--lesion boundary, as well as ENDO, a benign class whose morphology can resemble HSIL.

\textbf{CRIC Cervix.}
CRIC Cervix~\cite{rezende2021} contains six TBS categories---NILM,
ASC-US, LSIL, ASC-H, HSIL, and SCC---with 8,073, 1,730, and 1,731
annotated cells in the train, validation, and test splits, respectively.
Different scanning and staining conditions and a distinct demographic
distribution from RIVA provide a cross-dataset test of generalization.
NILM constitutes 58.8\% of the annotated cells, while ASC-US has only
91 validation and 91 test samples; its results should therefore be
interpreted with appropriate caution.

\textbf{Augmentations.}
Despite heavy class imbalance, no artificial augmentation or class
rebalancing was applied. The framework is therefore evaluated under the
original screening distributions, where benign cells dominate and
ambiguous lesion categories are naturally less frequent. This preserves
the native distribution encountered by an uncertainty-aware routing system while avoiding synthetic modification of the fine-grained nuclear and cytoplasmic morphology important for boundary-class discrimination.

\subsection{Metrics}
\label{sec:metrics}

We use \textbf{macro-F1} as the primary metric, together with
\textbf{lesion-F1} for the binary benign--lesion task and
\textbf{overall routing accuracy}. ROC-AUC and PR-AUC are also reported for the DST-fused output.

To assess whether uncertainty distinguishes reliable from unreliable
predictions, we report the \textit{uncertainty ratio} (UR):
\begin{equation}
\mathrm{UR} =
\frac{\mu_{m_U}(\mathrm{incorrect})}
     {\mu_{m_U}(\mathrm{correct})}.
\label{eq:ur}
\end{equation}
Thus, $\mathrm{UR}>1$ indicates greater mean uncertainty mass for
incorrect predictions. We further report performance for
high-confidence samples ($m_U\leq0.40$) and the uncertain pool
($m_U>0.50$). \textbf{\textit{Silent wrong}} (SW) denotes the fraction
of samples with $m_U\leq0.40$ routed to the incorrect binary branch.

\subsection{Implementation Details}
\label{sec:impl}

All scale-specific MLPs are trained independently using AdamW with
learning rate $10^{-3}$, weight decay $10^{-4}$, and CosineAnnealingLR
over 20 epochs. Features are extracted using a frozen DINOv2 ViT-B/14
backbone~\cite{oquab2023dinov2}. The best checkpoint is selected using
validation macro-F1 computed from the ordinally supervised predictions.

\section{Results}
\label{sec:results}

\subsection{Classification and Fusion Results}
\label{sec:classification_results}

Table~\ref{tab:mainresults} summarizes the benign--lesion routing
performance of the complete framework. Performance is consistently
stronger on CRIC Cervix, while RIVA remains the more challenging
eight-class setting. We next isolate the contribution of the fusion
strategy.


\begin{table}[t]
\centering
\caption{Overall benign--lesion routing performance using fused DST evidence. PR-AUC is reported due to the naturally imbalanced screening distribution.}
\label{tab:mainresults}
\small
\setlength{\tabcolsep}{3pt}
\renewcommand{\arraystretch}{0.95}
\begin{tabular}{@{}l>{\columncolor{gray!18}}c>{\columncolor{gray!18}}c
@{\hspace{10pt}}l>{\columncolor{gray!18}}c>{\columncolor{gray!18}}c@{}}
\toprule
\textbf{Metric} & \textbf{RIVA} & \textbf{CRIC}
& \textbf{Metric} & \textbf{RIVA} & \textbf{CRIC} \\
\midrule
Accuracy        & 0.750 & \textbf{0.927}
& Benign-F1      & 0.808 & \textbf{0.939} \\
Macro-F1        & 0.726 & \textbf{0.924}
& Lesion-F1      & 0.644 & \textbf{0.909} \\
Macro-Precision & 0.721 & \textbf{0.927}
& ROC-AUC        & 0.814 & \textbf{0.971} \\
Macro-Recall    & 0.736 & \textbf{0.921}
& PR-AUC         & 0.709 & \textbf{0.969} \\
\bottomrule
\end{tabular}
\end{table}

Table~\ref{tab:fusion_cmp} directly compares the individual spatial
views and alternative late-fusion strategies using the same calibrated scale-specific predictions, with the image embeddings and
scale-specific MLPs held fixed across all fusion methods. Thus, differences in fused performance isolate the effect of the fusion strategy rather than the representation or preceding classifiers. Majority voting combines the discrete scale decisions, probability averaging averages their lesion probabilities, inverse-entropy fusion assigns greater weight to predictions with lower entropy, and the learned MLP provides a trainable fusion
baseline. DST provides the strongest fusion performance on both datasets, with the clearest separation appearing on samples where the spatial views disagree.

\begin{table*}[t]
\centering
\scriptsize
\setlength{\tabcolsep}{4pt}
\renewcommand{\arraystretch}{0.95}

\definecolor{accent}{HTML}{EEF3FB}
\definecolor{accenttext}{HTML}{0C447C}

\caption{Comparison of individual scales and late-fusion strategies using the same calibrated scale-specific predictions. Accuracy and Macro-F1 are reported separately. Conflict results are computed only on samples for which the scale predictions disagree.}
\label{tab:fusion_cmp}

\resizebox{0.90\textwidth}{!}{%
\begin{tabular}{
p{2mm}l
cc
>{\columncolor{gray!19}}c
>{\columncolor{gray!19}}c
cc
>{\columncolor{gray!19}}c
>{\columncolor{gray!19}}c
}
\toprule

& &
\multicolumn{4}{c}{\textbf{RIVA}} &
\multicolumn{4}{c}{\textbf{CRIC Cervix}} \\

\cmidrule(lr){3-6}
\cmidrule(lr){7-10}

& &
\multicolumn{2}{c}{\textbf{Overall}} &
\multicolumn{2}{c}{\cellcolor{gray!19}\textbf{Conflict}} &
\multicolumn{2}{c}{\textbf{Overall}} &
\multicolumn{2}{c}{\cellcolor{gray!19}\textbf{Conflict}} \\

\cmidrule(lr){3-4}
\cmidrule(lr){5-6}
\cmidrule(lr){7-8}
\cmidrule(lr){9-10}

& \textbf{Method}
& \textbf{Acc.} & \textbf{MF1}
& \cellcolor{gray!19}\textbf{Acc.}
& \cellcolor{gray!19}\textbf{MF1}
& \textbf{Acc.} & \textbf{MF1}
& \cellcolor{gray!19}\textbf{Acc.}
& \cellcolor{gray!19}\textbf{MF1} \\

\midrule

\multirow{3}{*}{\rotatebox[origin=c]{90}{\scriptsize scales}}
&
100px (nucleus)
& .7293 & .7112
& .5476 & .5467
& .9012 & .8982
& .5510 & .5510 \\

&
200px (cytoplasm)
& .7163 & .6947
& .5153 & .5120
& .9185 & .9159
& .6531 & .6529 \\

&
400px (neighbourhood)
& .7260 & .6895
& .5393 & .5022
& .9006 & .8974
& .5476 & .5475 \\

\cmidrule(lr){2-10}

\multirow{4}{*}{\rotatebox[origin=c]{90}{\scriptsize fusion}}
&
Majority vote
& .7210 & .7071
& .5822 & .5815
& .9053 & .9031
& .7109 & .7125 \\

&
Probability averaging
& .7295 & .7147
& .5894 & .5850
& .9073 & .9065
& .7139 & .7107 \\

&
Inverse-entropy fusion
& \textbf{.7449} & \textbf{.7202}
& \textbf{.5939} & \textbf{.5932}
& .9110 & .9109
& .7381 & .7381 \\

&
Learned MLP fusion
& .7409 & .7173
& .5763 & .5763
& \textbf{.9141} & \textbf{.9122}
& \textbf{.7399} & \textbf{.7385} \\

\midrule

\rowcolor{accent}
\multicolumn{2}{l}{\textcolor{accenttext}{\textbf{DST fusion (ours)}}}
&
\textcolor{accenttext}{\textbf{.7502}}
& \textcolor{accenttext}{\textbf{.7259}}
& \textcolor{accenttext}{\textbf{.6178}}
& \textcolor{accenttext}{\textbf{.6150}}
& \textcolor{accenttext}{\textbf{.9266}}
& \textcolor{accenttext}{\textbf{.9238}}
& \textcolor{accenttext}{\textbf{.7517}}
& \textcolor{accenttext}{\textbf{.7515}}
\\

\bottomrule
\end{tabular}%
}

\vspace{1pt}
\footnotesize
\textbf{Acc. = Accuracy; MF1 = Macro-F1.}
\end{table*}

The conflict results are particularly relevant to the proposed
formulation: DST retains its advantage when spatial views provide
inconsistent evidence or insufficient evidence for a confident decision, rather than only when their predictions agree.

As discussed in Section~\ref{sec:related_works}, existing results arise from heterogeneous experimental settings; a detailed comparative study is provided in the supplementary material. Hence, their results provide valuable context but not a controlled baseline for our setting. Our framework targets a different objective: uncertainty-aware benign--lesion triage with an explicit state for insufficient evidence, rather than direct multiclass classification. A controlled comparison requires fixing the representation, predictions, and task while varying only the fusion strategy, precisely what Table~\ref{tab:fusion_cmp} provides by evaluating the same calibrated scale-specific predictions under our DST formulation against established late-fusion alternatives.

\subsection{Uncertainty and Clinical Ambiguity}
\label{sec:clinical_alignment}

Table~\ref{tab:perclass} shows a clear class-dependent pattern in the
fused uncertainty. Endpoint categories such as NILM and SCC are routed
more reliably and with lower uncertainty, whereas categories near the
benign--lesion boundary are less consistently resolved.


\begin{table}[t]
\centering
\small
\setlength{\tabcolsep}{3.5pt}
\renewcommand{\arraystretch}{1.0}

\definecolor{highlight}{HTML}{DDF2FF}
\definecolor{summary}{HTML}{E5E5E5}

\caption{Per-class routing behaviour across datasets. Acc.: routing
accuracy; SW: silent wrong (\%); $\mu_U$: mean uncertainty mass;
UR: uncertainty ratio between incorrect and correct predictions.}
\label{tab:perclass}

\begin{tabular}{@{}lcccc@{\hspace{5pt}}cccc@{}}
\toprule
& \multicolumn{4}{c}{\textbf{RIVA}}
& \multicolumn{4}{c}{\textbf{CRIC Cervix}} \\
\cmidrule(lr){2-5}
\cmidrule(lr){6-9}

\textbf{Class}
& \textbf{Acc.} & \textbf{SW$\downarrow$} & $\boldsymbol{\mu_U}$
& $\boldsymbol{\mathrm{UR}\uparrow}$
& \textbf{Acc.} & \textbf{SW$\downarrow$} & $\boldsymbol{\mu_U}$
& $\boldsymbol{\mathrm{UR}\uparrow}$ \\

\midrule

NILM
& .913 & 1.4 & .091
& 4.13
& .943 & 1.7 & .074
& \cellcolor{highlight}\textbf{5.11} \\

ENDO
& .222 & 46.7 & .101 & .42
& -- & -- & -- & -- \\

INFL
& .742 & 5.5 & .218 & 1.69
& -- & -- & -- & -- \\

ASC-US
& .405
& \cellcolor{highlight}\textbf{40.5}
& .165
& \cellcolor{highlight}\textbf{.57}
& .637 & 9.9 & .215 & 1.66 \\

LSIL
& .519 & 12.6 & .278 & 1.10
& .863 & 4.4 & .138 & 1.05 \\

HSIL
& .739 & 3.7 & .176 & 2.27
& .992 & 0.0 & .022 & 3.14 \\

ASC-H
& .679 & 7.1 & .276 & 1.13
& .978 & 0.7 & .039 & 2.62 \\

SCC
& \cellcolor{highlight}\textbf{.959}
& \cellcolor{highlight}\textbf{0.9}
& .047
& 2.48
& \cellcolor{highlight}\textbf{1.000}
& \cellcolor{highlight}\textbf{0.0}
& .023
& \cellcolor{highlight}\textbf{8.59} \\

\midrule

\rowcolor{summary}
\textbf{Overall}
& \textbf{.742} & \textbf{5.8} & \textbf{.166} & \textbf{1.94}
& \textbf{.943} & \textbf{1.7} & \textbf{.074} & \textbf{4.13} \\

\bottomrule
\end{tabular}
\end{table}

The uncertainty ratio exceeds one overall on both datasets, indicating
that incorrect predictions tend to carry more uncertainty. ASC-US is
the notable exception: correctly routed samples can remain more
uncertain than incorrect ones, showing that the uncertainty measure
captures genuine ambiguity rather than serving only as an error score.
ENDO remains the main failure mode on RIVA, with poor routing accuracy
and a high silent-wrong rate. We have addressed this further in ~\ref{sec:limitations}


\begin{figure*}[t]
\centering
\begin{tabular}{cc}
\includegraphics[width=0.48\textwidth]{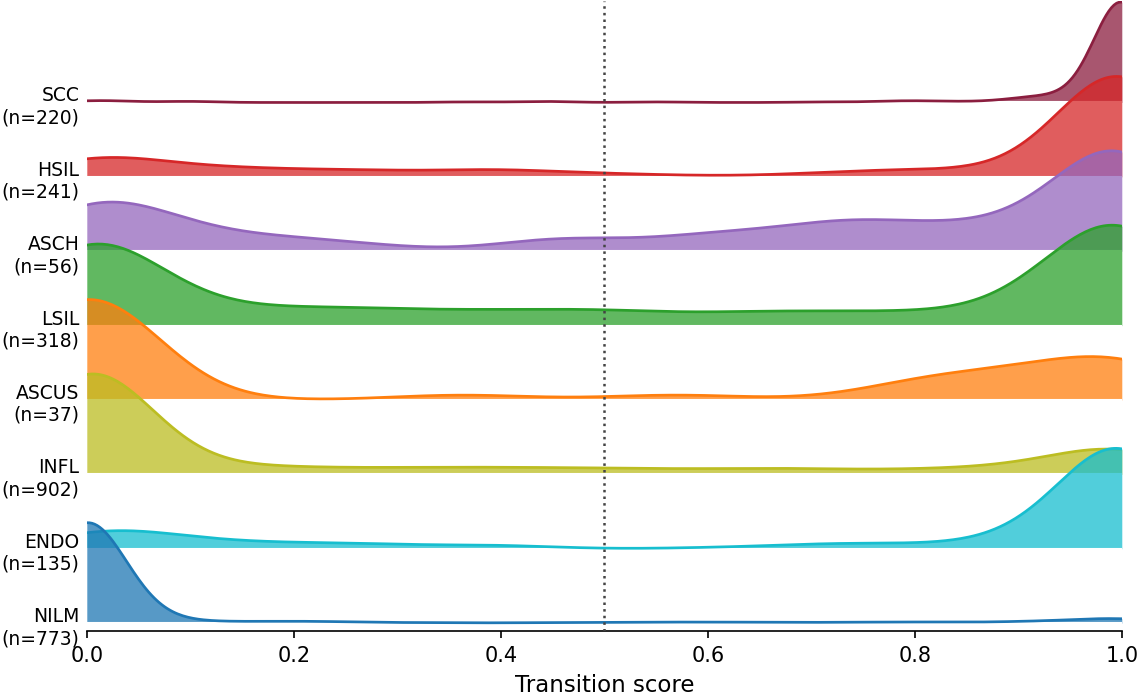} &
\includegraphics[width=0.48\textwidth]{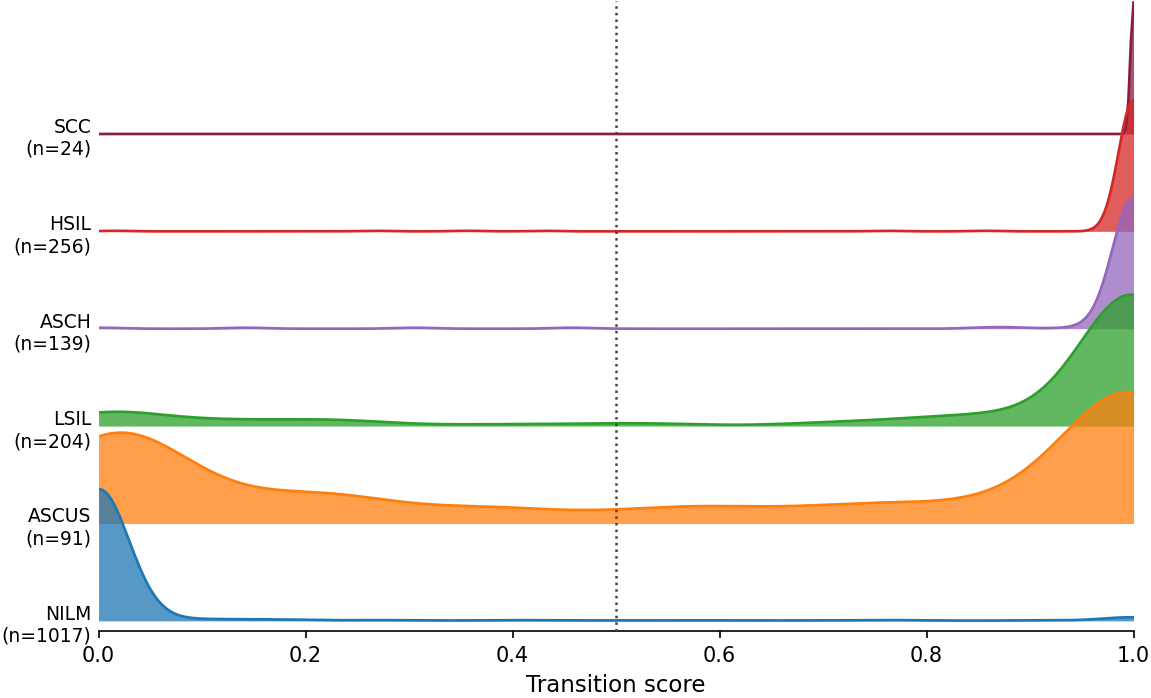} \\
(a) RIVA &
(b) CRIC Cervix \\
\end{tabular}
\caption{
Per-class distributions of fused lesion-transition scores across the
RIVA and CRIC Cervix test sets. The distributions retain the ordinal
progression from benign to malignant categories after multiscale
evidence fusion.
}
\label{fig:kde_progression}
\end{figure*}

Figure~\ref{fig:kde_progression} shows the distribution of fused
lesion-transition scores for each Bethesda category. NILM and SCC remain
concentrated near opposite ends of the score range, while boundary
categories such as ASC-US and LSIL occupy broader transitional regions.
This ordering is preserved across both datasets, showing that the ordinal
information learned before fusion remains encoded in the fused evidence,
despite the final routing framework operating on benign--lesion evidence. Representative multiscale routing cases are shown in Fig.~\ref{fig:dst-routing}.


\definecolor{cCorrect}{RGB}{46,160,67}

\definecolor{cBenign}{RGB}{31,119,180}
\definecolor{cLesion}{RGB}{214,39,40}
\definecolor{cUncertain}{RGB}{255,127,14}

\definecolor{cRouteBenign}{RGB}{76,175,80}
\definecolor{cRouteLesion}{RGB}{198,75,68}
\definecolor{cRouteUncertain}{RGB}{218,165,32}
\definecolor{cRouteAmbiguous}{RGB}{205,120,55}
\definecolor{cRouteDisagree}{RGB}{135,105,130}

\newcommand{\pbar}[3]{%
\begin{tikzpicture}[baseline=-0.45ex]
  \pgfmathsetmacro{\barlen}{0.82*#1}
  \draw[black!18,line width=2.1pt,line cap=round]
    (0,0) -- (0.82,0);
  \draw[#2,line width=2.1pt,line cap=round]
    (0,0) -- (\barlen,0);
  \fill[#2] (\barlen,0) circle (0.026);
  \pgfmathsetmacro{\threshpos}{0.82*#3}
  \draw[black,line width=0.45pt]
    (\threshpos,-0.075) -- (\threshpos,0.075);
  \node[font=\scriptsize,anchor=north]
    at (0,-0.065) {$#1$};
\end{tikzpicture}%
}

\newcommand{\dstbars}[3]{%
\begin{tikzpicture}[
  x=1cm,
  y=1cm,
  baseline=-0.45ex
]
  \pgfmathsetmacro{\bbar}{1.00*#1}
  \pgfmathsetmacro{\lbar}{1.00*#2}
  \pgfmathsetmacro{\ubar}{1.00*#3}

  \node[font=\scriptsize\bfseries,anchor=east]
    at (0.18,0.22) {B};
  \draw[black!18,line width=2pt,line cap=round]
    (0.25,0.22) -- (1.25,0.22);
  \draw[cBenign,line width=2pt,line cap=round]
    (0.25,0.22) -- ({0.25+\bbar},0.22);
  \node[font=\scriptsize,anchor=west]
    at (1.34,0.22) {$#1$};

  \node[font=\scriptsize\bfseries,anchor=east]
    at (0.18,0.00) {L};
  \draw[black!18,line width=2pt,line cap=round]
    (0.25,0.00) -- (1.25,0.00);
  \draw[cLesion,line width=2pt,line cap=round]
    (0.25,0.00) -- ({0.25+\lbar},0.00);
  \node[font=\scriptsize,anchor=west]
    at (1.34,0.00) {$#2$};

  \node[font=\scriptsize\bfseries,anchor=east]
    at (0.18,-0.22) {U};
  \draw[black!18,line width=2pt,line cap=round]
    (0.25,-0.22) -- (1.25,-0.22);
  \draw[cUncertain,line width=2pt,line cap=round]
    (0.25,-0.22) -- ({0.25+\ubar},-0.22);
  \node[font=\scriptsize,anchor=west]
    at (1.34,-0.22) {$#3$};
\end{tikzpicture}%
}

\newcommand{\votes}[2]{%
\begin{tabular}{@{}c@{}}
\scriptsize\textcolor{cLesion}{\textbf{#1 lesion}}\\[-1pt]
\scriptsize\textcolor{cBenign}{\textbf{#2 benign}}
\end{tabular}%
}

\newcommand{\route}[2]{%
\textcolor{#1}{\textbf{#2}}%
}

\begin{figure*}[t]
\centering
\caption{
Representative multiscale evidential routing cases from the RIVA and
CRIC Cervix test sets. Each row shows an actual test sample, its
scale-specific lesion probabilities and votes, the DST-fused evidence,
and the resulting routing action. The examples collectively illustrate
all five routing categories: confident benign, confident lesion, genuine
uncertain, ambiguous agreement, and disagreement resolved by DST.
}
\label{fig:dst-routing}

\resizebox{\textwidth}{!}{%
\begin{tabular}{@{}l l ccc c c l@{}}
\toprule
\textbf{Dataset}
& \textbf{True class / outcome}
& \textbf{100px}
& \textbf{200px}
& \textbf{400px}
& \textbf{Scale votes}
& \textbf{DST fused evidence}
& \textbf{Routing}
\\
\midrule

\multirow{3}{*}{\textbf{RIVA}}
&
LSIL: Lesion {\textcolor{cCorrect}{\checkmark}}
& \pbar{0.945}{cLesion}{0.456}
& \pbar{0.967}{cLesion}{0.447}
& \pbar{0.998}{cLesion}{0.496}
& \votes{3}{0}
& \dstbars{0.000}{0.997}{0.003}
& \route{cRouteLesion}{Confident lesion $\rightarrow$ lesion}
\\[4pt]

&
INFL: Benign {\textcolor{cCorrect}{\checkmark}}
& \pbar{0.233}{cBenign}{0.456}
& \pbar{0.605}{cLesion}{0.447}
& \pbar{0.357}{cBenign}{0.496}
& \votes{1}{2}
& \dstbars{0.314}{0.148}{0.538}
& \route{cRouteUncertain}{Genuine uncertain $\rightarrow$ further review}
\\[4pt]

&
LSIL: Lesion {\textcolor{cCorrect}{\checkmark}}
& \pbar{0.861}{cLesion}{0.456}
& \pbar{0.584}{cLesion}{0.447}
& \pbar{0.343}{cBenign}{0.496}
& \votes{2}{1}
& \dstbars{0.154}{0.370}{0.476}
& \route{cRouteDisagree}{Disagreement $\rightarrow$ DST resolves to lesion}
\\[5pt]

\midrule

\multirow{3}{*}{\textbf{CRIC Cervix}}
&
NILM: Benign {\textcolor{cCorrect}{\checkmark}}
& \pbar{0.000}{cBenign}{0.423}
& \pbar{0.008}{cBenign}{0.577}
& \pbar{0.016}{cBenign}{0.504}
& \votes{0}{3}
& \dstbars{0.987}{0.000}{0.013}
& \route{cRouteBenign}{Confident benign $\rightarrow$ benign}
\\[4pt]

&
NILM: Benign {\textcolor{cCorrect}{\checkmark}}
& \pbar{0.395}{cBenign}{0.423}
& \pbar{0.455}{cBenign}{0.577}
& \pbar{0.273}{cBenign}{0.504}
& \votes{0}{3}
& \dstbars{0.402}{0.000}{0.598}
& \route{cRouteAmbiguous}{Ambiguous (agree) $\rightarrow$ further review}
\\[4pt]

&
ASC-US: Lesion {\textcolor{cCorrect}{\checkmark}}
& \pbar{0.850}{cLesion}{0.423}
& \pbar{0.455}{cBenign}{0.577}
& \pbar{0.610}{cLesion}{0.504}
& \votes{2}{1}
& \dstbars{0.082}{0.350}{0.568}
& \route{cRouteUncertain}{Genuine uncertain $\rightarrow$ further review}
\\

\bottomrule
\end{tabular}%
}

\vspace{1pt}

\footnotesize
\textit{
\textbf{B, L, and U denote benign, lesion, and insufficient evidence,
respectively. Bar length denotes probability/evidence magnitude;
black vertical bars indicate the scale-specific decision thresholds.
All probabilities and fused masses are reported from the corresponding
test-set samples.}
}

\end{figure*}

\subsection{Ablation, Sensitivity, and Cross-Dataset Stability}
\label{sec:ablation}

Table~\ref{tab:design_transfer} separates the formulation into choices
that are shared across datasets, operating parameters that are
validation-adapted, and the single dataset-specific correction applied
to CRIC Cervix.


\begin{table}[t]
\centering
\small
\setlength{\tabcolsep}{3.5pt}
\renewcommand{\arraystretch}{0.95}
\caption{Cross-dataset status of principal design choices.}
\label{tab:design_transfer}
\begin{tabular}{@{}l l c@{}}
\toprule
\textbf{Component} & \textbf{Setting} & \textbf{Status} \\
\midrule
Multiscale crops
& 100/200/400 px
& Shared \\

Ordinal structure
& ASC-US $=0.60$
& Shared \\

Binary evidential frame
& $\Theta=\{B,L\}$
& Shared \\

Reliability discounting
& $\alpha\in[0.40,1.00]$
& Adapted \\

DST fusion
& Sequential
& Shared \\

BPA sharpness
& $\lambda=3$
& Shared \\

Uncertainty threshold
& $m_U^{\mathrm{high}}=0.50$
& Shared \\

Calibration
& Temperature / Isotonic
& Adapted \\

Scale thresholds
& $t_s$
& Adapted \\

CRIC NILM filter
& 2/3 lesion votes
& Dataset-specific \\
\bottomrule
\end{tabular}
\end{table}

The core evidential formulation transfers unchanged across datasets, while calibration and scale-specific thresholds are adapted to each dataset's validation distribution. The CRIC NILM filter is an additional dataset-specific refinement used to obtain a purer Genuine Uncertain pool, rather than a requirement for the framework to operate.

Sensitivity analysis shows stable behaviour across the main numerical
choices. The ASC-US anchor changes the degree of lesion commitment near
the boundary, with $0.60$ retaining stable behaviour across both
datasets. The reliability range primarily changes the relative
influence of the spatial sources, while $\lambda$ and
$m_U^{\mathrm{high}}$ mainly affect the size and composition of the
uncertain pool.

Finally, changing the order of sequential DST combination produces no
differences on either dataset, consistent with the commutative and
associative properties of Dempster's rule of combination. Additional ablations and sensitivity analyses of the evidential formulation, including the effects of fusion parameters and design choices, are provided in the supplementary material.

\section{Discussion}
\label{sec:discussion}

\textbf{Uncertainty as a first-class output.}
The uncertainty-stratified results (Table~\ref{tab:stratified}) show that $m_U$ is predictive rather than post-hoc: high-confidence samples outperform the overall average, while uncertain samples underperform it, concentrating uncertainty on diagnostically difficult cases. This identifies the uncertain pool as the natural target for further review.

\textbf{Biology-driven uncertainty structure.}
Per-class $\mu_U$ follows the ambiguity gradient of the Bethesda System, arising from the interaction of the soft ordinal targets and DST fusion. The ASCUS $\mu_U$-inversion is particularly informative: correctly predicted ASCUS samples remain more uncertain than confidently misclassified ones, indicating that correct recognition can retain genuine boundary ambiguity rather than confident assignment to either benign or lesion states.

\begin{table}[t]
\centering
\small
\setlength{\tabcolsep}{4pt}
\caption{
Uncertainty-stratified routing performance. High-confidence predictions
($m_U \leq 0.40$) substantially outperform uncertain predictions
($m_U > 0.50$), indicating that DST uncertainty mass is associated with
prediction reliability.
}
\label{tab:stratified}
\begin{tabular}{@{}lcccc@{}}
\toprule
\textbf{Stratum} &
\multicolumn{2}{c}{\textbf{RIVA}} &
\multicolumn{2}{c}{\textbf{CRIC}} \\
\cmidrule(lr){2-3}
\cmidrule(l){4-5}
&
\textbf{Macro-F1} &
\textbf{Acc.} &
\textbf{Macro-F1} &
\textbf{Acc.} \\
\midrule

All
& 0.7259 & 0.7502
& 0.9238 & 0.9266 \\

High confidence
& \textbf{0.7568} & \textbf{0.7830}
& \textbf{0.9403} & \textbf{0.9424} \\

Uncertain
& 0.5250 & 0.5345
& 0.5897 & 0.7500 \\

\bottomrule
\end{tabular}

\vspace{1mm}

{\scriptsize
\textbf{RIVA/CRIC sample counts are 2,682/1,731 for all samples,
2,276/1,649 for high-confidence samples, and
275/16 for uncertain samples.}
}
\end{table}

\section{Conclusion}
\label{sec:conclusion}

We presented a clinically grounded multiscale evidential framework for cervical cytology that combines nuclear, cellular, and neighbourhood evidence through calibrated Dempster--Shafer fusion. Across RIVA and CRIC Cervix, DST consistently improves over individual scales and alternative fusion strategies, with its strongest advantage occurring under inter-scale conflict. The fused evidence also preserves the ordinal progression of Bethesda categories, while incorrect predictions exhibit substantially higher uncertainty than correct predictions, particularly around ambiguous class boundaries. These results establish the value of explicitly representing inter-scale conflict and diagnostic ambiguity for more reliable cytology decisions and evidence-based routing.

\section{Limitations}
\label{sec:limitations}

\textbf{ENDO routing.}
ENDO is the clearest failure of the binary benign--lesion routing:
RIVA accuracy is 0.222, with 46.7\% silent-wrong rate and UR$=0.42$,
so incorrect predictions carry less uncertainty than correct ones.
Its benign morphology can resemble HSIL, exposing a limitation of the binary evidential frame; a richer hypothesis space for such morphological confounds remains an important direction for future work.

\textbf{CRIC uncertain pool.}
The genuine uncertain pool on CRIC contains only 16 test samples (0.9\%), with limited class representation; class-level routing statistics are therefore directional. The CRIC-specific NILM vote filter yields a purer pool, but its threshold must be determined empirically for new datasets.

{
    \small
    \bibliographystyle{ieeenat_fullname}
    \bibliography{main}
}

\end{document}